\documentclass[runningheads]{llncs}
\usepackage{graphicx}
\usepackage{epsfig}
\usepackage{times}
\usepackage{soul}
\usepackage{url}
\usepackage[utf8]{inputenc}
\usepackage[small]{caption}
\usepackage{graphicx}
\usepackage{hyperref}
\usepackage{booktabs}
\usepackage{amsfonts}       
\usepackage{nicefrac}       
\usepackage{microtype}      
\usepackage{amsmath,amssymb,amsfonts}
\usepackage{subfigure}
\usepackage{subfigmat}
\usepackage{tabularx}
\usepackage{bigstrut,multirow,rotating,threeparttable}
\usepackage{enumerate}
\usepackage{textcomp}
\usepackage{xcolor}
\usepackage{float}
\usepackage{stfloats}
\usepackage{pifont}
\usepackage{caption}
\usepackage{color}
\usepackage{colortbl}
\usepackage{wrapfig}
\usepackage{graphicx}
\usepackage{eucal}
\usepackage{bm}
\usepackage{bbm}
\usepackage{makecell}
\usepackage{algorithm,algorithmicx,algpseudocode}
\usepackage{graphicx}
\begin{document}
%
\title{Enhancing Zero-Shot Anomaly Detection: CLIP-SAM Collaboration with Cascaded Prompts}
%
%
\author{Yanning Hou\inst{1} \and Ke Xu$^{\ast}$\inst{1,2,3} \and Junfa Li\inst{1} \and Yanran Ruan\inst{1} \and Jianfeng Qiu\inst{1,2,3} }
 
\authorrunning{Yanning Hou and et al.}
%
\institute{
  School of Artificial Intelligence, Anhui University, Hefei, China
\and
Anhui Provincial Key Laboratory of Security Artificial Intelligence, Anhui University  \and
Anhui Provincial Engineering Research Center for Unmanned System and Intelligent Technology \\
\email{\{yanning\_hou,junfali,yanran\_ruan\}@stu.ahu.edu.cn}
\email{\{qiujianf,xuke\}@ahu.edu.cn}}

\maketitle              
\begin{abstract}
Recently, the powerful generalization ability exhibited by foundation models has brought forth new solutions for zero-shot anomaly segmentation tasks. However, guiding these foundation models correctly to address downstream tasks remains a challenge. This paper proposes a novel two-stage framework, for zero-shot anomaly segmentation tasks in industrial anomaly detection. This framework excellently leverages the powerful anomaly localization capability of CLIP and the boundary perception ability of SAM.
(1) To mitigate SAM's inclination towards object segmentation, we propose the Co-Feature Point Prompt Generation (PPG) module. This module collaboratively utilizes CLIP and SAM to generate positive and negative point prompts, guiding SAM to focus on segmenting anomalous regions rather than the entire object. (2) To further optimize SAM's segmentation results and mitigate rough boundaries and isolated noise, we introduce the Cascaded Prompts for SAM (CPS) module. This module employs hybrid prompts cascaded with a lightweight decoder of SAM, achieving precise segmentation of anomalous regions.
Across multiple datasets, consistent experimental validation demonstrates that our approach achieves state-of-the-art zero-shot anomaly segmentation results. Particularly noteworthy is our performance on the Visa dataset, where we outperform the state-of-the-art methods by 10.3\% and 7.7\% in terms of {$F_1$-max} and AP metrics, respectively.

\keywords{Anomaly Detection \and Zero-Shot \and CLIP \and SAM.}
\end{abstract}

\section{Introduction}
Zero-shot anomaly segmentation (ZSAS)~\cite{zsas1,zsas2,zsas3} is a crucial aspect of industrial anomaly detection, especially given the challenges posed by the scarcity of anomaly samples and the variability of anomaly types in real-world scenarios. Traditional approaches to anomaly segmentation, including self-supervised~\cite{selfsupervised1,selfsupervised2,selfsupervised3,selfsupervised4,selfsupervised5} and unsupervised~\cite{unsupervised1,unsupervised2,unsupervised3,unsupervised4} methods, have been extensively explored in previous research endeavors~\cite{pre_work1,pre_work2,pre_work3}. These approaches typically involve learning representations of normal samples during training and subsequently detecting anomalies by computing differences between test samples and the learned normal distribution. However, a significant drawback of these methods is the requirement for substantial amounts of data spanning diverse categories, which can be impractical for industries dealing with millions of products. Therefore, research on ZSAS is especially important for the industry.

The advent of foundational models such as CLIP~\cite{CLIP} and SAM~\cite{SAM} has revolutionized the field of zero-shot anomaly segmentation. These models leverage advanced techniques, to effectively identify and segment anomalies based on textual or positional prompts. By harnessing the capabilities of these foundational models, it becomes possible to achieve zero-shot anomaly segmentation without the need for extensive training data. This breakthrough not only enhances the feasibility of anomaly detection in data-scarce environments but also opens up new avenues for addressing anomaly detection challenges in various industrial applications.
In practical terms, leveraging CLIP~\cite{CLIP} and SAM~\cite{SAM} for zero-shot anomaly segmentation involves providing textual or positional prompts that guide the models to identify anomalies without prior training on specific anomaly types. For instance, in an industrial setting, textual descriptions or positional information related to product features can serve as prompts for the models to detect anomalies. This approach significantly reduces the dependence on labeled anomaly data and enables anomaly detection in diverse and dynamic industrial environments.

Many studies have already conducted zero-shot anomaly segmentation based on these foundational models, such as those based on CLIP~\cite{Anomalyclip,APRIL-GAN,SDP,WinCilp,Anovl}, SAM~\cite{SAA}, and CLIP\&SAM collaboration~\cite{CilpSAM}. The CLIP-based approach aligns text features with image features to achieve anomaly localization and segmentation, but it cannot effectively perceive anomaly boundaries. SAM-based methods utilize various prompts to guide localization, enabling effective perception of boundaries in anomalous regions. However, the prompt types are too fixed, primarily relying on bounding boxes, and the localization capabilities are severely limited. The collaborative approach of CLIP\&SAM~\cite{CilpSAM} suggests employing CLIP for localization and utilizing SAM for segmentation, showcasing robust anomaly perception and segmentation capabilities. However, existing CLIP\&SAM collaboration methods~\cite{CilpSAM} fail to fully leverage the respective abilities of CLIP~\cite{CLIP} and SAM~\cite{SAM}. Currently, the method exclusively depends on CLIP to directly supply point and bounding box prompts for SAM. While this strict prompt strategy prevents SAM from segmenting entire objects, it also limits SAM's ability to perceive boundaries, as segmentation is confined by bounding box prompts. Furthermore, in this procedure, CLIP~\cite{CLIP} and SAM~\cite{SAM} undertake entirely separate tasks, with the time-consuming SAM image encoder's features being used solely for segmentation boundaries. We consider this as wasteful.

Specifically, our approach involves the collaborative use of CLIP and SAM.  To fully capitalize on the respective capabilities of CLIP and SAM while preventing SAM from segmenting entire objects, we introduce the Co-Feature Point Prompt Generation (PPG) module. By integrating anomaly maps from CLIP and image features from the SAM image encoder, we generate positive and negative point prompts for SAM from two perspectives: extreme anomaly values in anomalous regions and similarity in surrounding areas. This encourages SAM to prioritize segmenting positive point features while disregarding negative ones, thereby effectively identifying anomalous regions. To provide effective prompts and constraints for SAM, leveraging its boundary perception capabilities, and mitigating issues such as incomplete segmentation, blurry boundaries, and isolated noise, we propose the Cascaded Prompts for SAM (CPS) module. Through cascaded mixed prompts, this module progressively strengthens constraints on SAM, accurately guiding SAM to fully segment anomalous regions. Our main contributions can be summarized as follows:

\begin{itemize}
\item[$\bullet$] We propose a novel framework  for zero-shot detection tasks, which involves collaborative use of CLIP and SAM to achieve precise segmentation of anomalous regions through their cooperation.
\item[$\bullet$] To effectively locate anomalies, we devised the PPG module, leveraging CLIP and SAM to provide more accurate positive and negative prompts by comprehensively considering anomaly values and feature similarity. This enhancement significantly improves the performance of zero-shot detection.
\item[$\bullet$] In order to fully leverage SAM's fine-grained segmentation capability and boundary perception ability, we innovatively introduced the CPS module, which employs cascaded operations to further enhance detection precision and robustness without requiring additional extensive computations.
\item[$\bullet$] Consistent experimentation across multiple datasets has validated that our approach achieves state-of-the-art zero-shot anomaly segmentation results. Particularly noteworthy is our performance on the Visa dataset, where we surpass the state-of-the-art methods by 10.3\% and 7.7\% in {$F_1$-max} and AP metrics, respectively.
\end{itemize}
\section{Related Work}

\subsection{Foundation Models}

Foundation models~\cite{foundation1,foundation2,foundation3,VLM1,VLM2,VLM3} show an impressive ability to solve diverse vision tasks in a zero-shot manner. CLIP ~\cite{CLIP} is the first model to be pre-trained on a web-scale dataset of image-text pairs. It focuses on aligning multi-modal features and possesses robust semantic understanding abilities for both language and vision, demonstrating unprecedented generality. SAM~\cite{SAM} demonstrates a powerful ability to extract high-quality object segmentation masks in the open world. It achieves this goal by effectively utilizing various prompts such as points, boxes, and rough masks, enabling it to accurately delineate object boundaries.

\subsection{Zero-shot Anomaly Segmentation}

The zero-shot anomaly segmentation task currently has three mainstream methods. 
The first method is based on CLIP~\cite{CLIP}. For example, the pioneering method, WinCLIP~\cite{WinCilp}, utilizes a sliding window approach to extract multi-scale features and aligns them with textual features. APRIL-GAN~\cite{APRIL-GAN} employs features from different hierarchical levels and further refines feature alignment using linear layers. AnomallyCLIP~\cite{Anomalyclip} proposes to enhance textual feature generalization, while SDP~\cite{SDP} addresses noise issues during the encoding process. CLIP-based methods have been relatively successful in addressing zero-shot anomaly classification problems. However, for zero-shot anomaly segmentation, most methods utilize patch-based and bilinear interpolation techniques to handle anomalous map, which often result in imperfect delineation of anomaly boundaries.
The second approach is based on SAM ~\cite{SAM}, for instance, SAA ~\cite{SAA}, which provides bounding box prompts to SAM through Grounding DINO ~\cite{GroundingDINO} to achieve anomaly segmentation. However, due to the limited localization capability of Grounding DINO ~\cite{GroundingDINO}, it cannot accurately identify anomalies. Moreover, SAM tends to segment objects, which can lead to segmenting entire objects instead of anomalous regions.
The third category of methods combines CLIP ~\cite{CLIP} with SAM ~\cite{SAM}. By utilizing CLIP for localization and providing prompts to SAM, zero-shot anomaly segmentation can be achieved. CilpSAM ~\cite{CilpSAM} is developed based on this concept. However, after obtaining localization information from CLIP, they directly feed both point prompts and bounding box prompts into SAM~\cite{SAM} through rough masks, restricting SAM's segmentation tendency to the bounding box. This greatly limits the boundary perception capability of SAM~\cite{SAM}  and overall anomaly segmentation ability. Moreover, the most time-consuming SAM~\cite{SAM} image encoder is only used as a decoder to obtain image features, failing to fully utilize its powerful feature extraction capability.

\subsection{Rethink the Roles of CLIP and SAM in Zero-Shot Anomaly Detection}

CLIP~\cite{CLIP} possesses strong capabilities in aligning images with text. Utilizing the Visual Transformer (ViT)~\cite{VIT} enables multi-level feature extraction, followed by further alignment of features using linear layers. These linear layers learn to map features from different levels into the same space, enhancing their consistency and comparability. Such alignment enhances the representational capacity of features, leading to improved accuracy and robustness in subsequent tasks, such as anomaly detection or classification. It greatly enhances CLIP's perceptual ability towards anomalies, achieving effective localization. Besides, SAM~\cite{SAM} defines a novel task of prompt-based segmentation, aiming to return a segmentation mask for any given prompt. SAM~\cite{SAM} is extensively pre-trained on 11 million images using 1 billion masks, endowing it with powerful generalization and boundary perception capabilities, enabling effective boundary segmentation given prompts. Its robust performance has been validated across multiple tasks~\cite{AboutSAM1,AboutSAM2,AboutSAM3}.

\begin{figure}[!htb]
    \centering
    \includegraphics[width=1\textwidth]{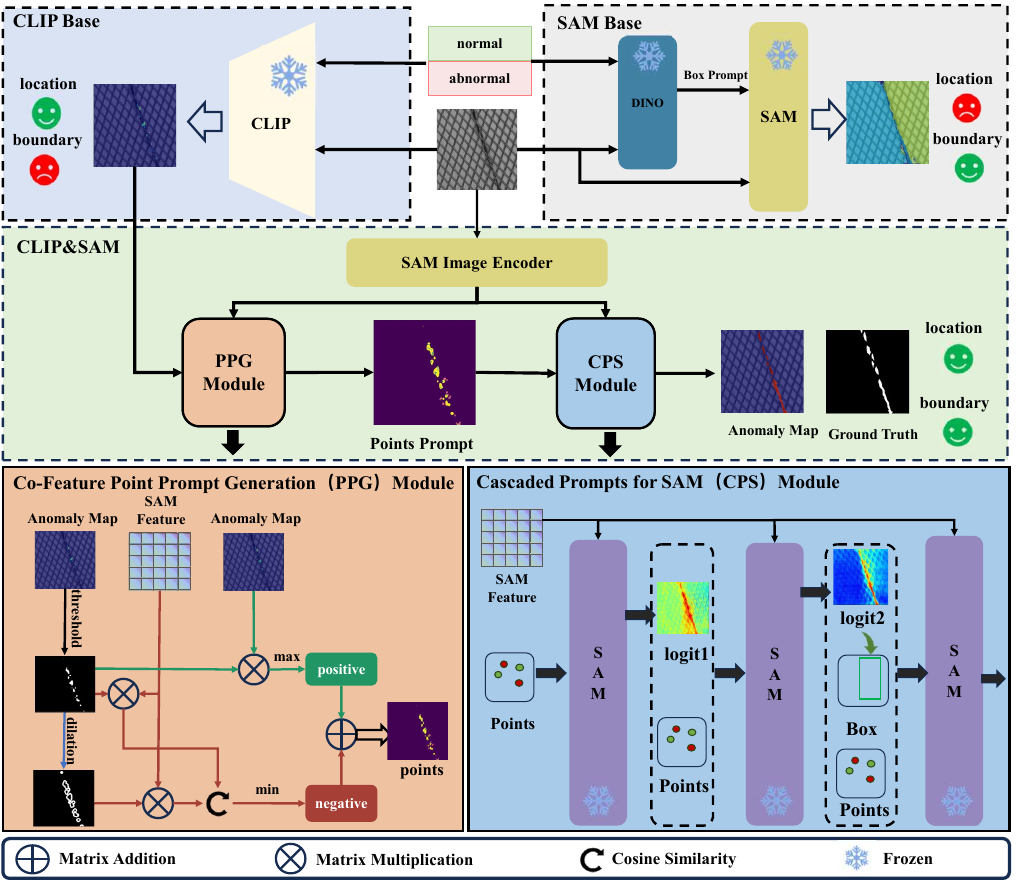} 
    \caption{The CLIP-based method aligns text and image features, enabling precise anomaly localization but struggles to fully segment the entire anomaly area and its boundaries. On the other hand, the SAM-based approach successfully segments boundaries but often confuses normal and abnormal regions. Our method integrates the strengths of these two foundational models. Through the Co-Feature Point Prompt Generation (PPG) module, we generate initial point prompts by leveraging CLIP~\cite{CLIP} and SAM~\cite{SAM}. Subsequently, via the Cascaded Prompts for SAM (CPS) module, we further refine the mask quality by cascading hybrid prompts for SAM~\cite{SAM}, ultimately achieving successful and accurate anomaly segmentation with our framework. }
    \label{fig:structure}
\end{figure}

In \autoref{fig:structure}, we can clearly observe the advantages and disadvantages of solely relying on these two approaches. Taking the thread grid as an example, the CLIP-based method accurately locates the anomaly region but fails to identify the entire anomalous area along with its boundaries. Conversely, the SAM-based approach precisely segments the image into two parts. However, due to its limited localization capability, the anomalous region is not accurately delineated.
In this paper, we propose a novel framework. Specifically, we utilize CLIP~\cite{CLIP} to identify extremely anomalous regions within anomaly images, which serve as prompts for SAM~\cite{SAM}.

\section{Methodology}
In this section, we provide a detailed explanation of the motivation and specifics of our approach. In Section 3.1, we collaborate CLIP and SAM to provide positive and negative point prompts for SAM, enabling anomaly localization. In Section 3.2, we cascade prompts to the mask decoder of SAM, allowing it to accurately segment abnormal boundaries comprehensively.

\subsection{CLIP\&SAM Co-Feature Point Prompt Generation}
\label{sec3.1}
For the anomaly map provided by CLIP~\cite{CLIP}  obtained through the threshold: we propose utilizing the Co-Feature Point Prompt Generation(PPG) module to provide positive and negative points prompts to SAM~\cite{SAM}, thereby guiding SAM~\cite{SAM} to accurately segment the entire anomalous region. 
\subsubsection{Localization of positive points}
After applying CLIP~\cite{CLIP}, we generated an anomaly map ($\mathrm{S}_{a}$) and identified regions of extreme anomaly using a threshold. We derived the anomaly map of the extremely anomalous regions ($\mathrm{R}_{a}$) by intersecting the extremely anomalous regions ($\mathrm{S}_{a}$) with the anomaly map ($\mathrm{Map}_{a}$). Subsequently, we selected the top $\mathrm k$ anomalous points based on their anomaly scores, considering them as positive points (spaced by 400 pixels).
\begin{equation}
\mathrm{R}_{a}=\mathrm{S}_{a} \otimes \mathrm{Map}_{a}  , 
\end{equation}

\begin{equation}
\mathrm{P}_{h}=\mathrm{Top}_{k}(\mathrm{R}_{a})  , 
\end{equation}
where $\otimes$ denotes element-wise multiplication. Here, Equation (1) represents the intersection operation between the anomaly map ($\mathrm{S}_{a}$) and the anomaly map of extremely anomalous regions ($Map_a$), resulting in the set of extremely anomalous regions ($\mathrm{R}_{a}$). Equation (2) denotes the selection of the top $\mathrm k$ anomalous points from $\mathrm{R}_{a}$, which are designated as positive points ($\mathrm{P}_{h}$).
\subsubsection{Localization of Negative points}
For SAM~\cite{SAM} handling both positive and negative prompts simultaneously, the selection of negative instances is particularly critical. If negative points are chosen solely based on global anomaly scores, they often represent background or regions far from the anomaly area. Such prompts may lead SAM to segment the entire object rather than focusing on the anomaly region, resulting in ineffective negative prompts. Therefore, we initially apply a dilation function to capture the surrounding regions of extreme anomaly areas($\mathrm{R}_{a}$) , generating negative prompts on these surrounding regions($\mathrm{N}_{a}$). This approach ensures that SAM directs its attention specifically towards the anomaly region, providing more effective negative prompts and enhancing segmentation accuracy.

\begin{equation}
\mathrm{N}_{a} =\mathrm{dilate}(\mathrm{S}_{a}) - \mathrm{S}_{a}, 
\end{equation}
Additionally, we utilize an image encoder to extract global features ($\mathrm{F}$). This encoder can be the frozen backbone network of SAM~\cite{SAM} or other pretrained visual models~\cite{DINO,DINOv2,resnet}. In our study, we default to using the SAM image encoder, which exhibits strong boundary perception. Moreover, the features extracted by this image encoder are also utilized for the SAM mask decoder. After obtaining global features, we employ spatial multiplication to compute local features of extreme anomaly regions ($\mathrm{F}_{a}$) and their surrounding areas ($\mathrm{F}_{n}$).
\begin{equation}
\mathrm{F}=\mathrm{Enc}_{I}({img}) ,
\end{equation}
\begin{equation}
\mathrm{F}_{a}=\mathrm{F} \otimes \mathrm{S}_{a}, \quad \mathrm{F}_{n}=\mathrm{F} \otimes \mathrm{N}_{a}, 
\end{equation}
Subsequently, we compute the cosine similarity between the local features($Map_s$) of these two parts and select the $k$ pixels with the lowest similarity as negative samples(spaced by 400 pixels).
\begin{equation}
\mathrm{Map_s}=Similarity(\mathrm{F}_{a}, \mathrm{F}_{n}) , 
\end{equation}
\begin{equation}
\mathrm{P}_{l}=\mathrm{Lowest}_{k}(\mathrm{Map_s}) , 
\end{equation} 
In this way, SAM would tend to segment the contiguous region surrounding the positive point, while discarding the negative one's on the image. Then, we combine the obtained positive ($P_h$) and negative ($P_l$) prompts together with the image features and seed them collectively into the decoder. Finally, we obtain the mask with the highest score.
\begin{equation}
\mathrm{P}=Contact(\mathrm{P}_{h}, \mathrm{P}_{l}),
\end{equation}
\begin{equation}
\mathrm{M}_{1},\mathrm{logit}_{1}=\mathrm{Dec}_{m}(\mathrm{F},P) , 
\end{equation}

\subsection{Cascaded Prompts for SAM}
\label{sec3.2}
With the aforementioned techniques, we obtain positive and negative points prompts for SAM~\cite{SAM}, along with the initial masks ($\mathrm{M}_{1}$) derived from these prompts and their corresponding logit($\mathrm{logit}_{1}$). 
Although the positive and negative point prompts effectively guide SAM~\cite{SAM} to segment positive features and discard negative ones, relying solely on these prompts, due to their granularity and sparsity, may result in the mask containing rough edges from the background and isolated noise points. To further refine the mask, we employ the Cascaded Prompts for SAM method. 

\subsubsection{Points+logit1}
SAM not only outputs segmentation masks but also generates low-resolution logit related to the segmentation. We utilize these logit as dense prompts fed back into SAM~\cite{SAM} because they are aligned with the spatial layout of the image, allowing for refinement of the mask edges and achieving clearer boundaries. By combining point prompts and dense logit($logit_1$) prompts, we obtain the segmentation mask($M_2$) for the second step. 
\begin{equation}
\mathrm{M}_{2},\mathrm{logit}_{2}=\mathrm{Dec}_{m}(\mathrm{F},Contact( \mathrm{P},\mathrm{logit}_{1})) , 
\end{equation}

\subsubsection{Points+box+logit2}
Anomalies typically occur in specific regions and are not widespread. Through the combination of point prompts and dense logit prompts, we can segment the majority of anomalies. However, there may still be rough noise present at spatially distant locations. Therefore, precise localization of anomaly positions is crucial. We utilize the highest-scored mask output from the previous SAM level to obtain its positional information and derive a bounding box. This information, combined with the point prompts and logit($\mathrm{logit}_{2}$) from the previous level, forms multi-type prompts fed into SAM~\cite{SAM} to obtain the refined final mask. 
\begin{equation}
\mathrm{box}=\mathrm{F}_{location}(\mathrm{M}_{2}) , 
\end{equation}
\begin{equation}
\mathrm{M}_{3}=\mathrm{Dec}_{m}(\mathrm{F},Contact( \mathrm{P},\mathrm{box},\mathrm{logit}_{2})) , 
\end{equation}
Due to our requirement of a lightweight decoder for iterative refinement, rather than a large-scale image encoder, the post-processing efficiency is high, with only an additional 100 milliseconds overhead. However, segmentation results show a significant improvement, with clear distinctions made for abnormal boundaries.
\section{Experiments}
In this section, we conducted extensive experiments to validate the effectiveness of our approach. In Section 4.1, we provide detailed insights into our experimental setup. In Section 4.2, we evaluate the performance of our method on various downstream tasks (MVTec-AD~\cite{MVTec} and VisA~\cite{visa}) and compare it with various ZSAS methods, accompanied by visualizations. Finally, in Section 4.3, we perform ablation studies to examine the impact of different designs on our method.

\subsection{Experimental Setup}
We conducted a series of experiments to evaluate the anomaly segmentation performance of our method in a zero-shot setting, covering the latest and challenging industrial anomaly segmentation benchmarks we focused on. We also conducted extensive ablation studies to validate the individual effectiveness of each component proposed by us.
\subsubsection{Datasets and Metrics}
We assessed the performance using two publicly available datasets: MVTec-AD~\cite{MVTec} and VisA~\cite{visa}. They contain high-resolution images of common objects with the full pixel-level annotations.
We conducted a fair and comprehensive comparison with existing zero-shot anomaly detection and segmentation (ZSAS) methods using widely adopted metrics, namely AUROC,  {$F_1$-max} and AP. Specifically, AUROC reflects the model's ability to differentiate between classes at various threshold levels.  {$F_1$-max} represents the harmonic mean of precision and recall at the optimal threshold, implying the model's accuracy and coverage. AP quantifies the accuracy of the model at different recall levels. Higher values of these metrics indicate better performance of the evaluation methodology.
\subsubsection{Implementation details}
In our experiments, we employed the pre-trained ViT-L-14-336 model released by OpenAI as the CLIP encoder, which consists of 24 Transformer layers. We extracted image patch embeddings after each stage of the image encoder (i.e., layers 6, 12, 18, and 24), which were used to train linear layers separately. We followed the same training setup as existing zero-shot anomaly segmentation~\cite{APRIL-GAN} studies. Specifically, the model was initially trained on the MVTec-AD~\cite{MVTec} dataset and then tested on the VisA~\cite{visa} dataset, and vice versa. We employed the Adam optimizer with a fixed learning rate of 1e-3. For the standard VisA dataset, training was conducted on a single GPU (NVIDIA GeForce RTX 3090) with a batch size of 16 for 3 epochs. As for the MVTec-AD dataset, the training duration was set to 15 epochs. For SAM, we use the ViT-H pre-trained model.
\subsection{Comparison with the State-of-the-Art}
In this section, we conducted an efficacy assessment of our proposed method, for zero-shot segmentation on the MVTec-AD~\cite{MVTec} and VisA~\cite{visa} datasets. \autoref{tab:sota} presents a comprehensive comparison between our proposed method, and state-of-the-art Zero-Shot Anomaly Segmentation (ZSAS) methods across various datasets and metrics. The conclusion drawn is that our proposed method, outperforms existing state-of-the-art methods across all {$F_1$-max} and AP metrics.

On the VisA~\cite{visa} dataset, our method achieved improvements of 10.3\% and 7.7\% in {$F_1$-max} and AP metrics, respectively. On the MVTec-AD~\cite{MVTec} dataset, we observed enhancements of 2.1\% and 1.1\%, respectively. However, in terms of the AUROC metric, we were respectively lower than the state-of-the-art methods by 3.0\% and 0.8\%. This discrepancy is attributed to our reliance on the SAM segmentation results as the primary reference, resulting in a wider span between anomalies and consequently poorer performance on the AUROC metric.
\begin{table}[!htb]
  \centering
    \begin{tabular}{cccccccccc}
    \hline
    \multicolumn{2}{c}{\multirow{2}[4]{*}{\textbf{Base model}}} & \multicolumn{2}{c}{\multirow{2}[4]{*}{\textbf{Method}}} & \multicolumn{3}{c}{\textbf{MVTec-AD}} & \multicolumn{3}{c}{\textbf{VisA}} \bigstrut\\
\cline{5-10}    \multicolumn{2}{c}{} & \multicolumn{2}{c}{} & \textbf{AUROC} & \textbf{$F_1$-max} & \textbf{AP} & \textbf{AUROC} & \textbf{$F_1$-max} & \textbf{AP} \bigstrut\\
    \hline
    \multicolumn{2}{c}{\multirow{5}[2]{*}{\makecell[c]{CLIP-based \\ Approaches}}} & \multicolumn{2}{c}{WinCLIP~\cite{WinCilp}} & 85.1  & 31.7  & -     & 79.6  & 14.8  & - \bigstrut[t]\\
    \multicolumn{2}{c}{} & \multicolumn{2}{c}{APRIL-GAN~\cite{APRIL-GAN}} & 87.6  & 43.3  & 40.8  & 94.2  & 32.3  & 25.7  \\
    \multicolumn{2}{c}{} & \multicolumn{2}{c}{SDP~\cite{SDP}} & 88.7  & 35.3  & 28.5  & 84.1  & 16.0  & 9.6  \\
    \multicolumn{2}{c}{} & \multicolumn{2}{c}{SDP+~\cite{SDP}} & \underline{91.2}  & 41.9  & 39.4  & 94.8  & 26.5  & 20.3  \\
    \multicolumn{2}{c}{} & \multicolumn{2}{c}{AnomalyCLIP~\cite{Anomalyclip}} & 91.1  & 39.1  & 34.5  & \underline{95.5}  & 28.3  & 21.3  \bigstrut[b]\\
    \hline
    \multicolumn{2}{c}{\multirow{2}[2]{*}{\makecell[c]{SAM-based \\ Approaches}}} & \multicolumn{2}{c}{SAA~\cite{SAA}} & 67.7  & 23.8  & 15.2  & 83.7  & 12.8  & 5.5  \bigstrut[t]\\
    \multicolumn{2}{c}{} & \multicolumn{2}{c}{SAA+~\cite{SAA}} & 73.2  & 37.8  & 28.8  & 74.0  & 27.1  & 22.4  \bigstrut[b]\\
    \hline
    \multicolumn{2}{c}{\multirow{2}[2]{*}{CLIP\&SAM}} & \multicolumn{2}{c}{ClipSAM~\cite{CilpSAM}} & \textbf{92.3 } & \underline{47.8}  & \underline{45.9}  & \textbf{95.6 } & \underline{33.1}  & \underline{26.0}  \bigstrut[t]\\
    \multicolumn{2}{c}{} & \multicolumn{2}{c}{\textbf{Ours}} & 89.5  & \textbf{48.8 } & \textbf{46.4 } & 94.8  & \textbf{36.5 } & \textbf{28.0 } \bigstrut[b]\\
    \hline
    \end{tabular}
  \vspace{0.70em}
  \caption{Performance comparison of SOTA approaches on the MVTec-AD~\cite{MVTec} and VisA~\cite{visa} datasets. Evaluation metrics include AUROC, {$F_1$-max} and AP. Bold indicates the best performance and underline indicates the runner-up.}
   \label{tab:sota}
\end{table}%

  \begin{figure}[!htb]
    \centering
    \includegraphics[width=0.89\textwidth]{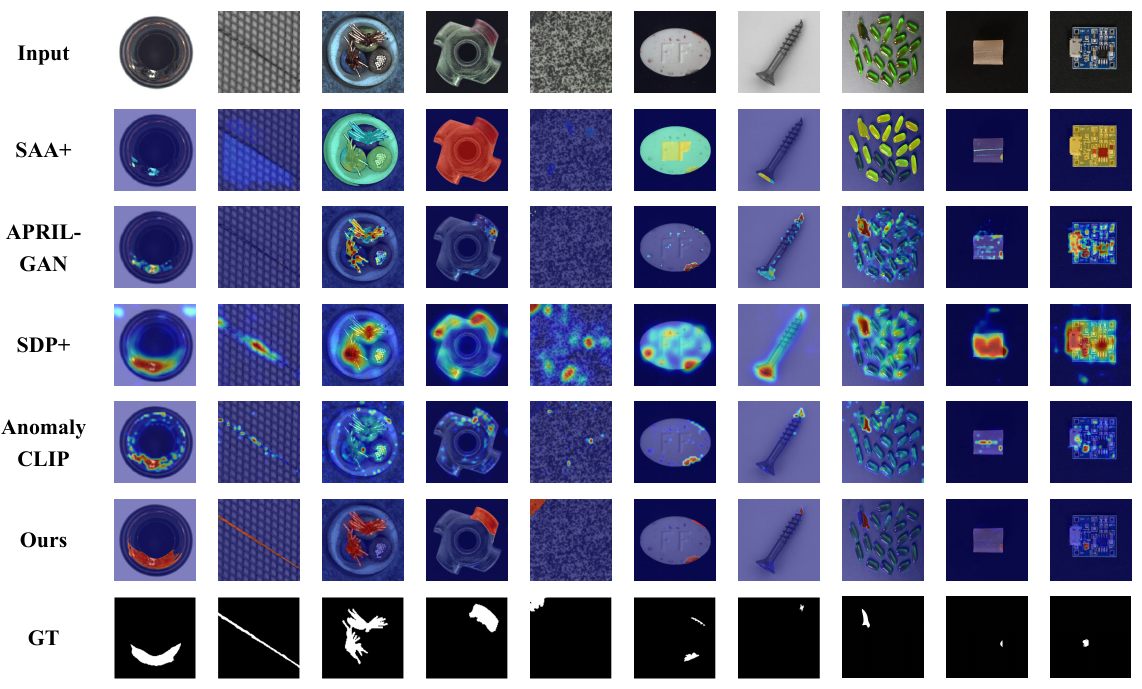} 
    \caption{Comparison of visualization results among SAA+~\cite{SAA}, APRIL-GAN~\cite{APRIL-GAN}, SDP+~\cite{SDP},  Anomaly-CLIP~\cite{Anomalyclip} and ours on the MVTec-AD~\cite{MVTec} dataset and VisA~\cite{visa} dataset.}
    \label{fig:sota}
\end{figure}

In Figure \ref{fig:sota}, we provide visualizations of some Zero-Shot Anomaly Segmentation (ZSAS) results to further demonstrate the effectiveness of the proposed method. For comparison, we also show the corresponding image results of SAA+~\cite{SAA}, APRIL-GAN~\cite{APRIL-GAN}, SDP+~\cite{SDP}, and Anomaly-CLIP~\cite{Anomalyclip}. It can be observed that the CLIP-based method performs well in anomaly localization. However, aligning text features with image features makes it difficult to locate boundaries, resulting in a considerable amount of noise  problem. Analyzing the results of SAA+~\cite{SAA}, it can be seen that the SAM-based method effectively identifies boundaries but lacks sufficient localization capability within the anomaly regions, leading to frequent misclassification of normal regions. Compared to these methods, our approach achieves superior anomaly region localization and segmentation, demonstrating stronger performance.

\subsection{Ablation Studies}
In this section, we conducted a series of ablation studies on the MVTec-AD dataset to further explore the impact of different components and experimental settings on the results of the proposed framework.

\subsubsection{Effect of dilation function kernel and kernel shape}
In our experiments, the  Co-Feature Point Prompt Generation(PPG) module is utilized to provide initial points prompts and serves as the foundation for the entire framework. The core of the PPG module lies in the utilization of dilation function, making the selection of dilation function parameters particularly crucial. Different kernel shapes and sizes can significantly impact the subsequent point prompts locations. Therefore, when designing the PPG module, careful consideration of the parameter settings of the dilation function is necessary to ensure it can provide accurate and effective initial points prompts, thereby laying a solid foundation for the operation of the entire framework. \autoref{tab:shape_size} displays the outcomes of ablation experiments involving nuclei of elliptical, rectangular, and cross shapes, with respective sizes of 20, 25, and 30. It's evident that employing an elliptical kernel shape with a size of (25, 25) achieves optimal results.

\begin{table}[!htb]
    \centering
    \begin{tabular}{c|c|c|c|c}
    \hline
      shape & size & AUROC & {$F_1$-max} & AP\\
      \hline
      cross & (20,20) &\textbf {89.5} & 46.8 & 44.1\\
      cross & (25,25) & 89.2 & 46.5 & 44.1\\
      cross & (30,30) & 89.3 & 46.3 & 44.5\\
      \hline
      rectangle & (20,20) & \textbf{89.5} & \underline{47.7} & \underline{45.6}\\
      rectangle & (25,25) & \underline{89.4} & 46.9 & 43.9\\
      rectangle & (30,30) & 89.0 & 45.0 & 42.2\\
      \hline
      ellipse & (20,20) & 89.1 & 47.2 & 45.3\\
      ellipse & (25,25) & \textbf{89.5} & \textbf{48.8} & \textbf{46.4}\\
      ellipse & (30.30) & 89.4 & 46.9 & 44.5\\
      \hline
    \end{tabular}
    \vspace{0.70em}
    \caption{The ablation study on different dilation function kernel shapes and sizes.}
    \label{tab:shape_size}
\end{table}

\subsubsection{Effect of cascade prompts}
\begin{figure}[!htb]
    \centering
    \includegraphics[width=0.67\textwidth]{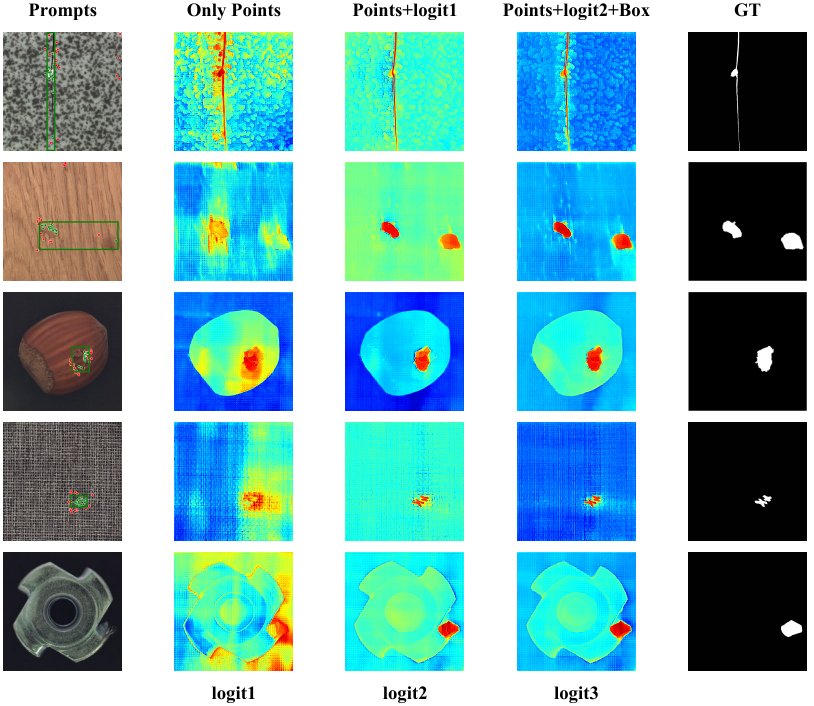} 
    \caption{Visualizations of SAM segmentation guided by the CPS module. When using point prompt Visualizations of SAM segmentation guided by the CPS module alone, the boundaries can be extremely blurry. With the addition of secondary points prompts and logit1, the delineation of abnormal boundaries becomes much clearer, although noise issues may persist. Upon introducing box prompt, the segmentation of boundaries can be achieved nearly perfectly.}
    \label{fig:step}
\end{figure}

The Cascaded Prompts for SAM (CPS) module, which cascades SAM three times in total, is now under discussion. We calculate the results for the first, second, and third stages separately in \autoref{tab:step}. After incorporating point prompts and logit1 at the second cascade level, the AUROC decreased by 0.6, while {$F_1$-max} increased by 4.3, and AP increased by 5.6. Finally, upon introducing the box prompt, the AUROC increased by 1.4, {$F_1$-max} increased by 2, and AP increased by 1.6, achieving optimal performance. we also providing partial image visualizations in \autoref{fig:step}. It's evident from the visualizations that after processing with the CPS module, rough boundaries and isolated noise points are greatly removed. This indicates that the CPS module offers a highly efficient and straightforward way of utilizing SAM.

\begin{table}[!htb]
    \centering
    \begin{tabular}{c|c|c|c}
    \toprule
         Cascaded & AUROC & {$F_1$-max} & AP  \\
         \midrule
         only points & 88.7 & 42.5 & 39.2\\
         points+logit1 & 88.1 & 46.8 & 44.8\\
         points+box+logit2 & \textbf{89.5} & \textbf{48.8} & \textbf{46.4}\\
         \bottomrule
    \end{tabular}
    \vspace{0.70em}
    \caption{The cascaded step ablation study on the MVTec-AD dataset. Results from the three-step cascade demonstrate, with bold indicating the best performance.}
    \label{tab:step}
\end{table}

\section{Conclusion}
We propose a novel collaborative framework between CLIP and SAM to address the zero-shot anomaly segmentation problem. To fully leverage the functionalities of these two base models, we introduce two modules. One is the PPG module, which combines the capabilities provided by CLIP and SAM to jointly determine initial point cues. The other is the CPS module, which further optimizes SAM segmentation by cascading blended type cues. Experiments demonstrate that our approach exploits the characteristics of different base models, offering new directions for improving ZSAS. While our method showcases robust zero-shot anomaly segmentation capabilities, the use of two models raises concerns regarding slower inference times. In future work, we will continue to explore how to efficiently and lightweightly integrate the advantages of different models to enhance anomaly segmentation capabilities.

\section*{Acknowledgments}
This work was supported in part by the National Natural Science Foundation of China (Grant no. 62206003) and The Anhui Provincial Natural Science Foundation (Grant no. 2308085MF201) and The Key Program of Natural Science Project of Educational Commission of Anhui Province (Grant no. KJ2021A0048 and KJ2021A0634).

\bibliographystyle{splncs04}
\bibliography{ref}

\end{document}